\documentclass{article}
\usepackage{iclr2027_conference,times}
\usepackage{graphicx}
\usepackage{booktabs}
\usepackage{multirow}
\usepackage{xcolor}
\usepackage{tabularx} 
\usepackage{capt-of}  
\usepackage{colortbl}
\usepackage{amsmath}
\usepackage{amssymb}
\usepackage{algorithm}
\usepackage{algorithmic}
\usepackage{wrapfig}
\usepackage[hyphens]{url}
\newcommand{\meansd}[2]{\ensuremath{#1_{\pm #2}}}
\definecolor{CalibNavy}{HTML}{17324D}
\definecolor{CalibTeal}{HTML}{2A7F83}
\definecolor{CalibAmber}{HTML}{D28A2D}
\definecolor{CalibBlue}{HTML}{3F6FA8}
\definecolor{CalibInk}{HTML}{263746}
\definecolor{CalibPaleBlue}{HTML}{EAF2F8}
\definecolor{CalibPaleGray}{HTML}{F1F3F5}
\definecolor{CalibPaleAmber}{HTML}{FFF3DF}
\usepackage{hyperref}
\title{CALIBUDGET: Calibration-Guided Source Allocation for Fixed-Budget Mixed-Reasoning Adaptation}
\author{%
   Yupeng Chang\textsuperscript{1}, Yuan Wu\textsuperscript{1,2}\thanks{Corresponding author} \\
   \textsuperscript{1}School of Artificial Intelligence, Jilin University\\
   \textsuperscript{2}Key Laboratory of Symbolic Computation and Knowledge Engineering, Jilin University\\
   \texttt{changyp23@mails.jlu.edu.cn, yuanwu@jlu.edu.cn} \\
}
\iclrfinaltrue

\begin{document}

\maketitle
\lhead{}

\begin{abstract}
Fixed-budget adaptation from heterogeneous data sources requires deciding not only how much data to use, but how much exposure each source receives. Size-proportional rules can crowd out small sources, whereas difficulty-only rules can chase noisy estimates or allocate residual budget to nearly saturated pools. We introduce CALIBUDGET, a floor-protected, reliability-aware integer allocator that treats source exposure as an explicit adaptation variable. From small train-internal calibration splits, it combines model need, post-floor availability, and bootstrap stability, then produces exact capacity-respecting quotas without changing the model, objective, or total budget. In a controlled setting combining mathematical and commonsense data, CALIBUDGET improves CommonAvg, FragileAvg, and MacroAvg over validation-error-with-floor, the strongest matched comparator, in all three paired LLaMA-2-7B LoRA+ runs. The respective mean gains are 0.56, 0.46, and 0.41 percentage points (pp). Overall increases by 0.18 pp, whereas MathAvg decreases by 0.20 pp, exposing a coverage--retention boundary rather than a uniform gain. CALIBUDGET changes only 1.14--1.42\% of the source budget but improves performance in 15 of 24 comparisons across commonsense tasks and seeds. These results suggest that small changes in source quotas can matter; example-level selection can then determine which examples fill each quota.
\end{abstract}

\section{Introduction}
Modern language-model adaptation relies on heterogeneous collections whose sources differ in size, supervision style, answer format, difficulty, and relevance to the capabilities being acquired or retained \citep{wei2021finetuned,sanh2021multitask,ouyang2022training,chung2024scaling,wang2022super}. In many adaptation rounds, the eligible pool exceeds what can be processed because of annotation, licensing, privacy, access, or compute constraints. Under such a cap, total data volume is fixed; the unresolved decision is how that budget should be distributed across sources. This exposure decision can matter even when the model, loss, and optimizer are unchanged, because aggregate accuracy can conceal underexposure of small, difficult, or format-distinct groups.

We study this setting as \emph{fixed-budget multi-source adaptation}. A heterogeneous candidate pool is partitioned into predefined sources, a companion training component is held constant, and the remaining budget must be converted into exact integer source quotas. Standard policies expose complementary failure modes. Pooled or size-proportional sampling favors large sources. Equal allocation protects coverage but ignores differences in model need and usable capacity. Difficulty-only allocation responds to model error, yet can overreact to noisy source-level estimates or prioritize sources with little residual capacity. A practical allocator must therefore reconcile \emph{need}, \emph{availability}, and \emph{estimation reliability} while satisfying hard budget, capacity, and minimum-exposure constraints.

Existing work optimizes data composition through continuous mixture weights, online sampling, proxy objectives, or example-level ranking \citep{xie2023doremi,xie2023data,xia2024less,liu2024makes,lin2024not}. These mechanisms are valuable, but they do not by themselves specify an exact feasible quota vector under source capacities and coverage requirements. We isolate this preceding decision: given predefined sources and an exact total budget, how many examples should be drawn from each source? This source-level allocation layer is orthogonal to within-source quality or diversity selection, and isolating it makes the effect of exposure directly measurable and auditable.

\begin{figure*}[!t]
\centering
\includegraphics[width=0.99\textwidth]{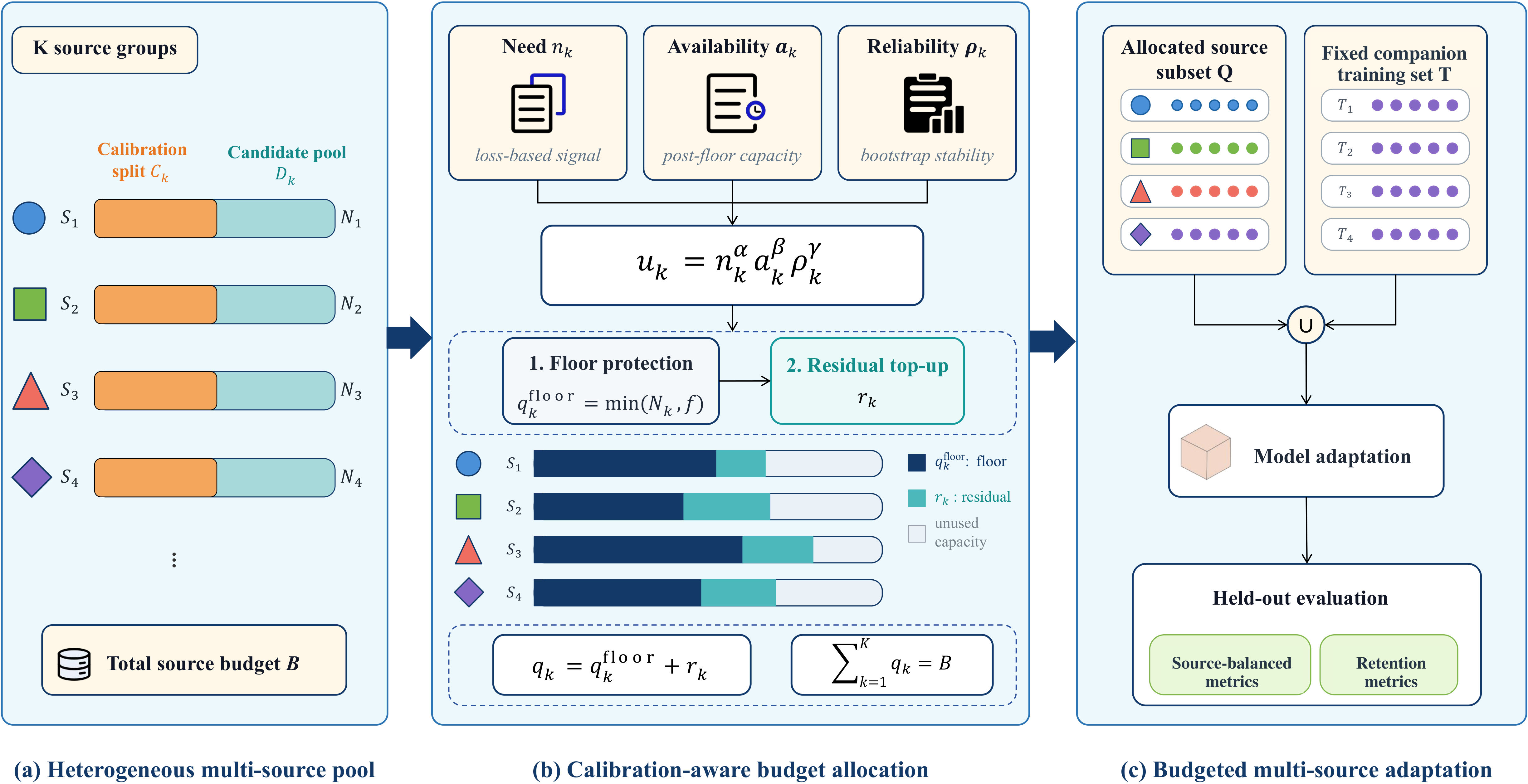}
\caption{Overview of CALIBUDGET. A heterogeneous multi-source pool is split into train-internal calibration data and candidate pools. Floor protection guarantees minimum exposure; post-floor availability, source need, and estimate reliability define the utility used for residual top-up, producing capacity-respecting integer quotas under budget $B$. The allocated source subset is combined with a fixed companion training set and adapted under an otherwise unchanged training protocol.}
\label{fig:overview}
\end{figure*}

We introduce \textbf{CALIBUDGET}, a calibration-guided integer allocator for fixed-budget multi-source adaptation. Figure~\ref{fig:overview} summarizes the design. CALIBUDGET reserves a small train-internal calibration split from each source and estimates three explicit signals: current model need, residual candidate availability, and the bootstrap stability of the need estimate. A floor first guarantees feasible minimum exposure; a utility-weighted residual stage then refines the allocation, followed by capacity-aware rounding that satisfies the budget exactly. The method neither encodes math- or commonsense-specific semantics nor changes the training objective. Its output is an inspectable quota vector determined by source statistics and feasibility constraints.

We evaluate this allocation layer in a controlled mixed-reasoning instantiation. A mathematical reasoning training subset is held fixed, while a capped budget is allocated across heterogeneous commonsense sources spanning yes/no, physical, social, adversarial-completion, and science reasoning. All paired comparisons use the same backbone, LoRA+ configuration, optimizer, schedule, parser, and random seed \citep{yu2023metamath,hayou2024lora+}. Across three LLaMA-2-7B seeds, CALIBUDGET improves CommonAvg, the prespecified fragile-task aggregate FragileAvg, and MacroAvg over the strongest matched validation-error-with-floor baseline in every seed, with mean gains of 0.56, 0.46, and 0.41 pp. Overall increases by 0.18 pp, whereas MathAvg decreases by 0.20 pp; we therefore characterize the result as a coverage-oriented trade-off, not a uniform improvement. The allocation changes only 1.14--1.42\% of the 10,000-example source budget yet improves 15 of 24 comparisons across commonsense tasks and seeds, showing that small, structured quota corrections can affect downstream task outcomes under this controlled setting. Our contributions are threefold:
\begin{itemize}
    \item We formulate capped heterogeneous adaptation as an exact source-quota problem, separating the choice of total data volume from the distribution of a fixed budget and distinguishing quota construction from continuous mixture weighting and example ranking.
    \item We develop CALIBUDGET, an auditable allocator that decomposes source exposure into a policy-controlled floor and a reliability-aware residual correction, while enforcing source capacities and exact integer feasibility.
    \item Under a strictly matched mixed-reasoning protocol, we show a consistent allocation-sensitivity pattern on source-balanced metrics and characterize its retention trade-off, mechanism diagnostics, operating-point sensitivity, and cross-backbone boundary.
\end{itemize}

\section{Method}

\subsection{Problem Formulation}

CALIBUDGET addresses fixed-budget multi-source adaptation. Let $D_s=\bigcup_{k=1}^{K}D_k$ be a heterogeneous candidate pool partitioned into $K$ predefined sources by dataset identity or metadata; each source is one allocation group. Each source provides a disjoint train-internal calibration split $C_k$. Let $\widetilde{T}$ denote a fixed companion training set included unchanged across methods. Given a pretrained model $p_{\theta_0}$, the allocator must select exactly $B$ examples from $D_s$ before learning an update $\Delta\theta$.

The selected multi-source subset is $Q=\bigcup_{k=1}^{K}Q_k$, where $Q_k\subseteq D_k$ and $|Q|=B$. The final adaptation set is
\begin{equation}
\widetilde{D}=\widetilde{T}\cup Q.
\end{equation}
Across paired comparisons, the model, adaptation backend, optimizer, schedule, evaluation pipeline, and random seed are fixed. The methodological object is therefore the quota vector $(q_1,\ldots,q_K)$ with $q_k=|Q_k|$: methods differ only in how they distribute the source budget $B$.

We instantiate this formulation in mixed-reasoning adaptation: $\widetilde{T}$ is a fixed mathematical training subset, and $D_s$ is the union of commonsense sources. The allocator itself uses only source-level calibration statistics and capacity constraints, not math- or commonsense-specific semantics.

\subsection{Calibration-Aware Source Utility}

CALIBUDGET weights residual allocation using three source-level factors: current model need, post-floor availability, and the stability of the need estimate. Using the disjoint calibration split $C_k$, we define
\begin{equation}
 u_k = n_k^{\alpha} a_k^{\beta} \rho_k^{\gamma},
\end{equation}
where $n_k$ is calibration need, $a_k$ is a transformed residual-availability signal, and $\rho_k$ is estimate reliability.

For calibration example $i\in C_k$, let
\begin{equation}
 z_i = \ell_i + \mathbf{1}\{\widehat{y}_i \neq y_i\},
\end{equation}
where $\ell_i$ is the mean per-token teacher-forced response NLL and $\widehat{y}_i$ is the greedily decoded answer under the fixed task parser. The indicator adds a unit penalty for a parser-level answer error, so $z_i$ combines graded model fit with discrete answer correctness. Let $m_k=|C_k|^{-1}\sum_{i\in C_k}z_i$, and let $\sigma_k$ be the standard deviation of 200 bootstrap means from $C_k$~\citep{efron1992bootstrap}. With a small numerical constant $\epsilon$, the frozen implementation uses
\begin{equation}
n_k = \frac{\max(m_k,\epsilon)}
{K^{-1}\sum_j \max(m_j,\epsilon)},\quad
\rho_k = (1+\sigma_k)^{-1},\quad
a_k = \sqrt{\max\!\left(N_k-\min(N_k,f),\epsilon\right)}.
\end{equation}
Each calibration split contains 100 held-out training examples, and $(\alpha,\beta,\gamma)=(1,0.5,1)$ is frozen before confirmatory runs. Because $a_k$ already applies a square-root transform, raw residual capacity enters the utility with exponent $1/4$. This deliberately tempers pool-size dominance while reducing preference for nearly exhausted sources.

The three factors address different failure modes. Proportional allocation reflects source size but not model need; validation-error allocation captures need but treats finite calibration estimates as exact and ignores residual capacity; equal allocation protects exposure but suppresses informative source differences. CALIBUDGET combines these signals only after a shared coverage floor has been assigned. Here, ``calibration'' refers to estimating allocation statistics on held-out training data, not to post-hoc probability calibration.

The mechanism is intentionally source-level rather than example-level. Ranking and influence methods answer which examples are preferable within a candidate set and usually require proxy or gradient computation. CALIBUDGET instead asks how many examples each source should contribute. Uniform within-source sampling isolates that decision; quality, diversity, or influence selectors can be composed within the resulting quotas.

\subsection{Floor-Protected Residual Allocation}

Let $N_k=|D_k|$ be the candidate size of source $k$. CALIBUDGET first assigns every feasible source a floor quota
\begin{equation}
 q_k^{\mathrm{floor}}=\min(N_k,f),
\end{equation}
where $f$ is chosen such that $\sum_{k} q_k^{\mathrm{floor}}\leq B$. The floor encodes a minimum-coverage requirement and prevents small or format-distinct sources from being crowded out. The residual budget is
\begin{equation}
 R=B-\sum_{k=1}^{K} q_k^{\mathrm{floor}}.
\end{equation}
Let $c_k=\max(N_k-q_k^{\mathrm{floor}},0)$ denote post-floor residual capacity. CALIBUDGET distributes $R$ over active sources according to normalized utility:
\begin{equation}
 \widetilde{r}_k = R\frac{u_k}{\sum_{j:c_j>0}u_j}.
\end{equation}
These residual shares are real-valued. Capacity-aware largest-remainder rounding converts them to integer allocations, enforces $0\le r_k\le c_k$, and iteratively redistributes overflow to sources with remaining capacity. The final source quota is
\begin{equation}
 q_k=q_k^{\mathrm{floor}}+r_k,
 \qquad \sum_{k=1}^{K} q_k=B.
\end{equation}
After quotas are fixed, we sample $q_k$ examples uniformly without replacement from each source and set $Q_k$ to the selected subset.

\paragraph{Allocation guarantee.}
If $\sum_k \min(N_k,f)\leq B\leq \sum_k N_k$, capped largest-remainder redistribution returns integer quotas satisfying
\begin{equation}
 \min(N_k,f)\le q_k\le N_k,
 \qquad \sum_k q_k=B.
\end{equation}
Thus every feasible source receives its prescribed minimum exposure and the source budget is met exactly. The guarantee concerns feasibility and coverage only; it does not imply that downstream performance is monotone in quota or that the resulting vector is globally optimal.

This floor--residual decomposition separates \emph{policy} from \emph{preference}. The floor determines the admissible coverage prior, whereas the calibrated utility refines only the remaining budget. If the floor is too small, high-error sources can dominate and small sources can be under-covered; if it is too large, the procedure approaches equal allocation and leaves little scope for model-dependent refinement. In our experimental instantiation, $B=10{,}000$ and $f=1150$, so the floor assigns 9,069 examples and leaves $R=931$ for utility-based correction. This conservative operating point tests whether calibration adds value beyond a dominant shared coverage prior. A prespecified $f=950$ audit evaluates sensitivity without retuning.

\begin{algorithm}[t]
\caption{CALIBUDGET}
\begin{algorithmic}[1]
\REQUIRE Candidate sources $\{D_k\}_{k=1}^{K}$, calibration splits $\{C_k\}_{k=1}^{K}$, budget $B$, floor $f$
\FOR{$k=1,\ldots,K$}
\STATE $q_k^{\mathrm{floor}}\leftarrow \min(|D_k|,f)$; $c_k\leftarrow |D_k|-q_k^{\mathrm{floor}}$
\STATE Estimate $n_k$ and $\rho_k$ from $C_k$; set $a_k\leftarrow\sqrt{\max(c_k,\epsilon)}$
\STATE $u_k\leftarrow n_k^{\alpha}a_k^{\beta}\rho_k^{\gamma}$
\ENDFOR
\STATE $R\leftarrow B-\sum_k q_k^{\mathrm{floor}}$
\STATE Set $\widetilde r_k=Ru_k/\sum_{j:c_j>0}u_j$ for active sources
\STATE Apply capacity-aware largest-remainder rounding, $0\le r_k\le c_k$
\STATE Sample $q_k=q_k^{\mathrm{floor}}+r_k$ examples from each $D_k$
\RETURN $Q=\bigcup_k Q_k$ with $\sum_k q_k=B$
\end{algorithmic}
\end{algorithm}

\subsection{Parameter-Efficient Adaptation}

After constructing $Q$, we train on $\widetilde{D}=\widetilde{T}\cup Q$ with LoRA+. For a pretrained weight matrix $W$, LoRA learns
\begin{equation}
 W' = W+\Delta W,
 \qquad \Delta W=\frac{s}{r}B_{\!L}A_{\!L},
\end{equation}
where $A_{\!L}$ and $B_{\!L}$ are rank-$r$ factors and $s$ is the scaling coefficient. LoRA+ assigns different learning rates to the two factors. We update attention and MLP projections and optimize the standard autoregressive objective:
\begin{equation}
\mathcal{L}(\Delta\theta)=
-\mathbb{E}_{(x,y)\sim \widetilde{D}}
\sum_{t=1}^{|y|}\log p_{\theta_0+\Delta\theta}(y_t\mid x,y_{<t}).
\end{equation}
No source-specific loss reweighting is introduced: once examples are selected, all are trained identically and source exposure is controlled only by the quotas. LoRA+ provides the controlled adaptation backend in our experiments, but CALIBUDGET does not use its internal parameterization; compatibility with other training regimes is conceptually direct but remains to be validated empirically.

\begin{table*}[!t]
\centering
\footnotesize
\setlength{\tabcolsep}{3.7pt}
\renewcommand{\arraystretch}{1.05}
\begin{tabular*}{\textwidth}{@{\extracolsep{\fill}}llccccc@{}}
\toprule
& & \multicolumn{1}{c}{Retention} & \multicolumn{2}{c}{Source-balanced} & \multicolumn{2}{c}{Aggregate} \\
\cmidrule(lr){3-3}\cmidrule(lr){4-5}\cmidrule(lr){6-7}
Method & Allocation signal & Math (\%) & Common (\%) & Fragile (\%) & Macro (\%) & Overall (\%) \\
\midrule
Pooled uniform & pooled random & 30.43 & 69.98 & 67.57 & 62.07 & 50.21 \\
Proportional & source size & 30.72 & 69.48 & 66.84 & 61.72 & 50.10 \\
Equal & equal source quota & 30.51 & 71.09 & 71.86 & 62.97 & 50.80 \\
Floor-Sqrt & floor + $\sqrt{\text{residual size}}$ & 30.47 & 71.03 & 71.19 & 62.92 & 50.75 \\
\rowcolor{CalibPaleGray}
Val-error+floor & floor + calibration error & \textbf{30.91} & \underline{71.31} & \underline{72.15} & \underline{63.23} & \underline{51.11} \\
\rowcolor{CalibPaleBlue}
\textbf{CALIBUDGET} & floor + $n a^{1/2}\rho$ & 30.72 & \textbf{71.87} & \textbf{72.62} & \textbf{63.64} & \textbf{51.29} \\
\bottomrule
\end{tabular*}
\caption{Three-seed mean performance (\%) on LLaMA-2-7B under matched data budgets, adaptation backend, and evaluation. Bold denotes the best descriptive mean; underlining marks the strongest matched comparator when it is runner-up.}
\label{tab:main}
\end{table*}

\section{Experiments}

We organize the evaluation around one confirmatory paired comparison, followed by mechanism diagnostics and transfer audits. The secondary analyses probe mechanism and boundary conditions; none is used to select or revise the frozen method.

\subsection{Experimental Protocol}

\paragraph{Training and budgets.}
The confirmatory protocol uses LLaMA-2-7B~\citep{touvron2023llama} with LoRA+~\citep{hayou2024lora+} for three epochs: rank $r=128$, scaling $s=128$, zero adapter dropout, geometric-mean factor learning rate $2\times10^{-5}$ with ratio 8, effective batch size 32, bf16, 1,024-token context, cosine decay, 3\% warmup, and zero weight decay. Adapters cover attention and MLP projections. We evaluate the final checkpoint, with no test-based checkpoint or allocation selection. The fixed companion set contains 100,000 MetaMath examples~\citep{yu2023metamath}; the eight-source commonsense pool receives exactly $B=10{,}000$ examples. Only source quotas and the corresponding uniformly sampled examples vary across methods, isolating allocation from changes in model capacity, parameterization, optimization, or total supervision.

\paragraph{Evaluation and baselines.}
We evaluate MATH and GSM8K~\citep{hendrycks2021measuring,cobbe2021training} together with BoolQ, PIQA, Social IQa, HellaSwag, WinoGrande, ARC-Easy, ARC-Challenge, and OpenBookQA~\citep{clark2019boolq,bisk2020piqa,sap2019social,zellers2019hellaswag,sakaguchi2020winogrande,clark2018think,mihaylov2018can}. MathAvg averages the two mathematical tasks, CommonAvg averages the eight commonsense tasks, and MacroAvg averages all ten tasks; Overall gives equal weight to MathAvg and CommonAvg. FragileAvg is a prespecified coverage-sensitive diagnostic over Social IQa, ARC-Easy, ARC-Challenge, and OpenBookQA. Unweighted task means prevent large evaluation sets from silently dominating the aggregate and align evaluation with the source-coverage motivation. All methods share fixed task parsers and scorers. Baselines are pooled-uniform, proportional, equal, Floor-Sqrt, and validation-error-with-floor allocation, spanning pooled, size-driven, coverage-driven, and difficulty-driven quota rules. The last baseline shares CALIBUDGET's calibration split and floor but allocates the residual budget by calibration error alone, making it the closest matched comparator. DoReMi, LESS, and RHO-1~\citep{xie2023doremi,xia2024less,lin2024not} require proxy, gradient, or token-level decisions and are not direct integer-quota substitutions; we do not claim to outperform them, and they can be composed with source quotas.

\paragraph{Controls.}
The CALIBUDGET configuration was frozen before the confirmatory comparison. Calibration examples come only from the training pool and are excluded from both adaptation and evaluation. Paired runs match the model, LoRA+ backend, fixed companion subset, total source budget, parser, and random seed; primary results average seeds 42--44. Exploratory variants and cross-backbone extensions are reported only as diagnostics and do not revise the frozen method.

\subsection{Main Paired Comparison}

Table~\ref{tab:main} reports the confirmatory LLaMA-2-7B comparison. CALIBUDGET achieves the highest CommonAvg, FragileAvg, MacroAvg, and Overall, while validation-error-with-floor retains the highest MathAvg. Relative to this strongest matched comparator, CALIBUDGET gains $+0.56$, $+0.46$, $+0.41$, and $+0.18$ pp on the four former metrics and loses $0.20$ pp on MathAvg. The central result is therefore not a uniform Pareto improvement: CALIBUDGET consistently improves CommonAvg, FragileAvg, and MacroAvg across seeds, with a mean MathAvg trade-off.

\begin{wraptable}[10]{r}{0.54\textwidth}
\vspace{-10pt}
\centering
\footnotesize
\setlength{\tabcolsep}{3.3pt}
\begin{tabular}{lrrrrr}
\toprule
Seed & Math & Common & Fragile & Macro & Overall \\
\midrule
42 & $-1.12$ & $+0.49$ & $+0.59$ & $+0.16$ & $-0.32$ \\
43 & $+0.27$ & $+0.38$ & $+0.32$ & $+0.36$ & $+0.33$ \\
44 & $+0.26$ & $+0.81$ & $+0.48$ & $+0.70$ & $+0.53$ \\
\midrule
\rowcolor{CalibPaleBlue}
Mean & $-0.20$ & $+0.56$ & $+0.46$ & $+0.41$ & $+0.18$ \\
SD & $0.80$ & $0.22$ & $0.14$ & $0.27$ & $0.44$ \\
\bottomrule
\end{tabular}
\caption{Paired performance differences (CALIBUDGET minus Val-error+floor, pp). Differences are computed before display rounding; SD is across seeds.}
\label{tab:deltas}
\end{wraptable}

Table~\ref{tab:deltas} resolves the mean result by seed. CommonAvg, FragileAvg, and MacroAvg improve in all three paired runs. Overall declines only on seed 42, where the MathAvg reduction is also largest; seeds 43--44 improve on both MathAvg and Overall. With only three seeds, these patterns are descriptive rather than large-sample statistical evidence, but their paired consistency on these metrics supports the primary allocation-sensitivity claim.

\subsection{Mechanism and Selection Diagnostics}

\begin{wraptable}[17]{R}{0.54\textwidth}
\centering
\footnotesize
\setlength{\tabcolsep}{1.6pt}
\renewcommand{\arraystretch}{1.02}
\begin{tabular}{lccccc}
\toprule
Diagnostic & Math & Common & Fragile & Macro & Overall \\
\midrule
\multicolumn{6}{l}{\emph{Frozen references, seed 42}} \\
\rowcolor{CalibPaleBlue}
CALIBUDGET & 30.20 & 71.29 & 71.79 & 63.07 & 50.74 \\
\rowcolor{CalibPaleGray}
Val-error+floor & 31.32 & 70.80 & 71.20 & 62.91 & 51.06 \\
\addlinespace[1pt]
\multicolumn{6}{l}{\emph{Exploratory normalization variants, seed 42}} \\
Group-norm A & 30.85 & 71.46 & 72.06 & 63.34 & 51.16 \\
Group-norm B & 30.48 & 71.84 & 72.86 & 63.57 & 51.16 \\
Group-norm C & 30.81 & 71.22 & 71.49 & 63.14 & 51.02 \\
B + recovery & 30.63 & 71.49 & 72.84 & 63.32 & 51.06 \\
\addlinespace[1pt]
\multicolumn{6}{l}{\emph{Fixed-index stability check}} \\
\rowcolor{CalibPaleAmber}
B indices, seed 43 & 30.53 & 70.92 & 72.40 & 62.84 & 50.73 \\
\bottomrule
\end{tabular}
\caption{Exploratory seed-42 normalization variants and a fixed-index seed-43 check (\%). Internal variants A--C alter source normalization; recovery adds a short retention-oriented phase. None revises the frozen CALIBUDGET configuration.}
\label{tab:diagnostics}
\end{wraptable}

\paragraph{Reliability and the matched comparator.}
Removing $\rho_k$ reduces CommonAvg, FragileAvg, MacroAvg, and Overall by 0.52, 0.95, 0.44, and 0.33 pp, respectively; the full rule is better on the first three metrics in every seed. This pattern is consistent with the intended role of $\rho_k$: discounting source-need estimates that fluctuate under bootstrap resampling instead of treating finite calibration means as equally trustworthy. Because the utility terms interact through normalization and quota rounding, the ablation is evidence for the reliability-aware rule rather than an isolated causal estimate of $\rho_k$. The validation-error-with-floor baseline remains stringent: its higher MathAvg but lower CommonAvg, FragileAvg, and MacroAvg illustrate the central coverage--retention boundary.

\begin{figure*}[!t]
\centering
\includegraphics[width=0.98\textwidth]{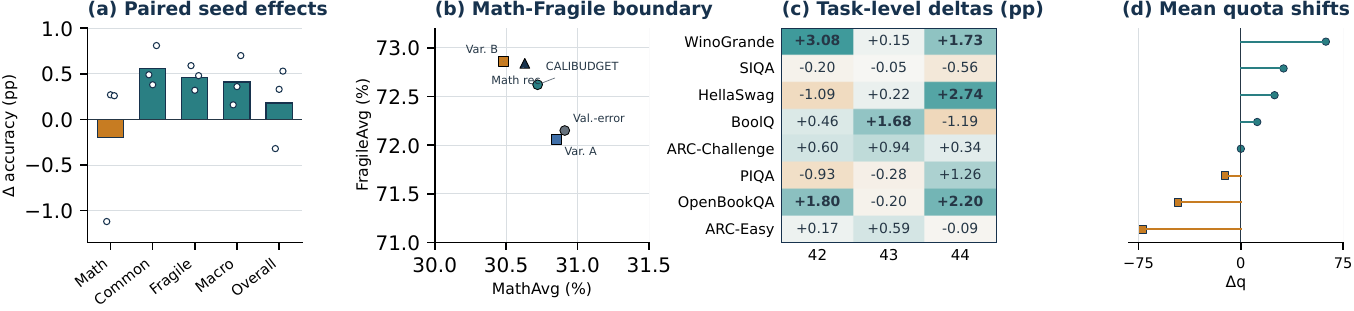}
\caption{Diagnostics under the frozen comparison: (a) paired seed effects relative to validation-error-with-floor; (b) the observed MathAvg--FragileAvg boundary, including exploratory variants; (c) task-level accuracy differences (pp); and (d) mean source-quota shifts under the 10,000-example budget. No diagnostic panel selects a method or hyperparameter.}
\label{fig:diagnostics}
\vspace{-5pt}
\end{figure*}

\paragraph{Guarding against seed-level selection.}
Table~\ref{tab:diagnostics} records exploratory normalization variants that improve seed-42 FragileAvg, but none was promoted. Variant B remains below validation-error-with-floor on MathAvg, the recovery phase does not remove the trade-off, and reusing the same selected indices does not reproduce the gain at seed 43. Reporting these negative checks prevents a favorable single-seed variant from replacing the prespecified rule post hoc.
\vspace{-5pt}
\subsection{Allocation Geometry and Task-Level Effects}
\vspace{-5pt}
Figure~\ref{fig:diagnostics} connects aggregate behavior to the underlying intervention. Relative to validation-error-with-floor, CALIBUDGET reallocates only 142, 135, and 114 of 10,000 source examples across seeds 42--44 (1.42\%, 1.35\%, and 1.14\%); no source changes by more than 76 examples. The performance differences therefore arise from a localized redistribution above a largely shared floor rather than from wholesale corpus replacement. Across the eight commonsense tasks, CALIBUDGET improves 15 of 24 task--seed comparisons, with recurrent gains on WinoGrande, OpenBookQA, ARC-Challenge, and HellaSwag. Quota and accuracy changes are nevertheless non-monotonic (Pearson $r=-0.005$; Spearman $\rho=-0.098$): some tasks improve after receiving fewer or unchanged examples. This behavior is plausible in multi-source instruction tuning because examples can transfer across skills and interact through shared parameters and the fixed mathematical component. The utility should therefore be interpreted as a joint allocation signal, not as a per-task treatment-effect estimator. This source-level role distinguishes CALIBUDGET from quality-, diversity-, and influence-based instance selectors~\citep{liu2024makes,bukharin2024data,xia2024less}, which can operate within allocated quotas.
\vspace{-5pt}
\subsection{Robustness and Transfer Audits}
\vspace{-5pt}
The matched LLaMA-2 study supports the primary claim. The following frozen-rule audits test whether the observed allocation effects transfer and are interpreted as boundary evidence rather than as opportunities to select a new method.
\vspace{-5pt}
\begin{table*}[!t]
\centering
\footnotesize
\setlength{\tabcolsep}{1.05pt}
\renewcommand{\arraystretch}{1.15}
\begin{tabular*}{\textwidth}{@{\extracolsep{\fill}}llccccc@{}}
\toprule
Model & Method & MathAvg (\%) & CommonAvg (\%) & FragileAvg (\%) & MacroAvg (\%) & Overall (\%) \\
\midrule
Qwen2.5-7B & Val-error+floor
& \meansd{65.77}{0.30} & \meansd{84.95}{0.29}
& \meansd{88.42}{0.37} & \meansd{81.12}{0.29}
& \meansd{75.36}{0.30} \\
& \textbf{CALIBUDGET}
& \meansd{66.13}{0.31} & \meansd{85.23}{0.21}
& \meansd{88.77}{0.32} & \meansd{81.41}{0.18}
& \meansd{75.68}{0.19} \\
\midrule
LLaMA-3.1-8B & Floor-Sqrt
& \meansd{60.77}{0.37} & \meansd{82.99}{0.17}
& \meansd{84.60}{0.42} & \meansd{78.54}{0.07}
& \meansd{71.88}{0.11} \\
& \textbf{CALIBUDGET}
& \meansd{60.72}{0.55} & \meansd{82.97}{0.19}
& \meansd{84.57}{0.26} & \meansd{78.52}{0.26}
& \meansd{71.84}{0.37} \\
\bottomrule
\end{tabular*}
\caption{Cross-backbone audits. Qwen2.5-7B~\citep{qwen2025qwen25technicalreport} averages three independently constructed adaptive allocations; LLaMA-3.1-8B~\citep{grattafiori2024llama} averages three training seeds while holding each method's seed-42 allocation fixed. Comparators are matched within each audit. Main values are mean accuracy (\%); subscripts report SD (pp).}
\label{tab:backbone}
\vspace{-5pt}
\end{table*}

\paragraph{Adaptive Qwen2.5 audit.}
The Qwen2.5 descriptive means favor CALIBUDGET on all five aggregates (Table~\ref{tab:backbone}). A paired bootstrap over the ten task-level scores yields deltas of $+0.35$ pp on MathAvg (95\% CI $[-0.38,+1.02]$), $+0.28$ pp on CommonAvg ($[-0.16,+0.69]$), $+0.29$ pp on MacroAvg ($[-0.17,+0.69]$), and $+0.32$ pp on Overall ($[-0.21,+0.76]$). Every interval crosses zero, and seed-level Overall differences are $-0.23$, $+0.61$, and $+0.57$ pp. Qwen2.5 therefore provides directionally consistent but seed-variable evidence, not backbone-invariant confirmation.

\paragraph{Fixed-allocation LLaMA-3.1 audit.}
The LLaMA-3.1 rows are effectively at parity. Unlike the adaptive Qwen2.5 audit, this experiment transfers each method's seed-42 allocation across three training seeds instead of recalibrating quotas for the new backbone. This near-parity result narrows the supported scope: because need is model-dependent, a quota vector estimated for one backbone should not be assumed to transfer without recalibration.

\paragraph{Modern-task and stability audits.}
On a frozen 320-prompt panel spanning BBH, GSM-Plus, HumanEval, IFEval, MATH-500, and MBPP~\citep{suzgun2023challenging,li2024gsm,chen2021evaluating,zhou2023instruction,lightman2024let,austin2021program}, strict ModernMacro deltas are $+0.61$ pp for Qwen2.5 (95\% CI $[-1.22,+2.34]$) and $+0.17$ pp for LLaMA-2 ($[-0.69,+1.04]$); all intervals cross zero; CodeMacro differences are zero. The panel broadens the evaluation but does not establish modern-task adaptation gains. By contrast, the allocation is stable across three independently constructed calibration splits: CALIBUDGET preserves quota ordering (Spearman $\rho=1.000$; Pearson $r\ge 0.9996$), with mean quota SD 1.47 and maximum range four examples, whereas validation-error-with-floor is less stable (mean SD 5.97, range 23, minimum $\rho=0.790$).

\paragraph{Floor sensitivity.}
Lowering $f$ from 1,150 to 950 expands the residual budget from 931 to 2,400 examples. At seed 42, CALIBUDGET retains relative gains over its matched comparator on CommonAvg ($+0.58$ pp), FragileAvg ($+0.57$ pp), MacroAvg ($+0.44$ pp), and Overall ($+0.23$ pp), but absolute performance falls below the prespecified continuation criterion. We stop before seeds 43--44. The relative ordering persists, but the lower-floor operating point is not sufficiently robust to promote, indicating that the floor is a consequential design constraint rather than an incidental constant.

\subsection{Interpretation and Limitations}

The matched LLaMA-2 LoRA+ experiments show that performance can depend on how a fixed source budget is allocated. Small, auditable quota changes improve CommonAvg, FragileAvg, and MacroAvg in all three seeds, while validation-error-with-floor retains slightly higher mean MathAvg. The reliability ablation supports the complete allocation rule, while the calibration-split audits show stable quotas. Transfer evidence is mixed: Qwen2.5 gains vary across seeds, LLaMA-3.1 performs similarly to its comparator under fixed allocations, and confidence intervals on the modern-task panel include zero. Lowering the floor preserves relative gains, but absolute performance falls below the same prespecified continuation criterion. Table~\ref{tab:claim-boundary} in Appendix~\ref{app:claim-boundaries} summarizes the supported claims and their limits.

Source capacities, the common floor, and the exact budget define feasible integer quotas; calibrated utility distributes the remaining budget. This design is useful when source groups are meaningful, candidate data exceed the training budget, minimum exposure is desirable, and train-internal calibration examples are available. It is less appropriate when all high-quality data can be used, source labels are arbitrary, or variation within sources dominates differences between them.

The mean MathAvg decrease accompanies gains in CommonAvg, FragileAvg, and MacroAvg, illustrating a trade-off between coverage and retention. Whether this trade-off is acceptable should be decided before evaluation. Source partitions and floors should be specified before test evaluation, rather than adjusted to improve downstream scores. Each quota can be traced to the floor, calibrated source statistics, remaining capacity, and deterministic rounding. Saving these quantities with selected indices and quota hashes allows exposure changes to be inspected. Capacity checks reject infeasible requests before training, and quality or diversity selection can be applied within each source.

The main study uses one setting combining mathematical and commonsense data, three seeds, predefined sources, and LoRA+; broader training domains and full-parameter tuning may yield different relationships between allocation and performance. CALIBUDGET does not model quality, diversity, contamination, or redundancy within sources, and the mathematical component remains fixed. The allocation guarantee covers exact integer feasibility and minimum coverage, not optimality or worst-group generalization. The need statistic uses teacher-forced NLL and parser-based correctness, while the bootstrap term measures sampling stability rather than full epistemic uncertainty. All primary allocations use only training-pool metadata and train-internal calibration examples; test results are not used to select quotas, hyperparameters, or checkpoints. We also retain configurations, seeds, parser versions, negative audit results, and per-task outputs to distinguish confirmatory comparisons from exploration. Future work could incorporate calibration uncertainty into quotas and jointly budget the mathematical and heterogeneous source components.

\section{Related Work}

\paragraph{Data mixtures and source-level budgeting.}
Data composition is a central design variable in pretraining and instruction tuning \citep{longpre2023flan,renduchintala2024smart,li2025data,shin2026dynamixsft}. DoReMi learns continuous domain-mixture weights through proxy training; RegMix and Data Mixing Laws predict promising mixtures from small-scale runs; and importance resampling estimates source or instance relevance to a target distribution \citep{xie2023doremi,liu2025regmix,ye2024data,xie2023data}. Recent instruction-tuning work further combines mixture design with submodular selection, scaling-law prediction, or dynamically updated sampling probabilities \citep{renduchintala2024smart,li2025data,shin2026dynamixsft}. At finer granularity, DEITA filters by complexity, quality, and diversity; QDIT studies quality--diversity trade-offs; LESS estimates target-specific gradient influence; and RHO-1 selects tokens online \citep{liu2024makes,bukharin2024data,xia2024less,lin2024not}. CALIBUDGET addresses an orthogonal discrete decision: before adaptation, it maps a finite budget to capacity-respecting integer quotas over predefined sources. Its floor guarantees feasible exposure and its rounding step satisfies the budget exactly; any example-level selector can subsequently rank candidates within each quota. Uniform within-source sampling is therefore a deliberate control that isolates source allocation.

\paragraph{Coverage, uncertainty, and adaptation backends.}
Group-robust learning improves minority-group performance by changing training weights or worst-group objectives \citep{sagawa2020investigation}; CALIBUDGET instead acts once before optimization, so its formal guarantee concerns integer feasibility and minimum exposure rather than worst-group risk. Its reliability term is a bootstrap stability discount on a source-level need statistic, not predictive probability calibration or a full uncertainty estimate under distribution shift \citep{efron1992bootstrap,guo2017calibration,snoek2019can}. Parameter-efficient methods update adapters, prefixes, or low-rank factors \citep{houlsby2019parameter,li2021prefix,hu2022lora,dettmers2023qlora}; AdaLoRA reallocates rank, LoRA+ changes factor-wise learning rates, and DoRA decomposes weight magnitude and direction \citep{zhang2023adalora,hayou2024lora+,liu2024dora}. These methods allocate parameter or optimization capacity, whereas CALIBUDGET allocates training examples and adds no PEFT module. We use LoRA+ as a controlled backend for mathematical and commonsense sources \citep{cobbe2021training,hendrycks2021measuring,yu2023metamath,clark2019boolq,bisk2020piqa,sap2019social,zellers2019hellaswag,sakaguchi2020winogrande,clark2018think,mihaylov2018can}; generalization beyond the tested models, protocol, and source partition remains an empirical question.

\section{Conclusion}
CALIBUDGET makes source exposure a first-class decision in fixed-budget adaptation. A minimum-coverage floor and reliability-aware residual correction yield exact, auditable quotas. Under matched LLaMA-2 LoRA+ controls, reallocating only 1.14--1.42\% of the source budget consistently improves CommonAvg, FragileAvg, and MacroAvg, while a small MathAvg decrease exposes the coverage--retention boundary. Transfer audits indicate that calibrated quotas remain model- and operating-point-dependent, positioning exact source budgeting as an adaptation control.

\clearpage
\section*{Reproducibility Statement}
Section~2 specifies the allocator and feasibility guarantee, and Section~3.1 reports the matched training and evaluation protocol. The appendix documents the allocation and audit implementation, deterministic integrity checks, local configuration boundary, and claim-specific audits. Primary allocations use only training-pool metadata and train-internal calibration examples; test results are not used to select quotas, hyperparameters, or checkpoints.

\section*{AI Use Disclosure}
Generative AI tools were used for manuscript drafting and language polishing, literature retrieval and reference-related tasks, minor figure-label editing, and feedback on experimental design and result interpretation. All AI-assisted outputs were reviewed by the authors and checked against the underlying sources, code, and experimental records where applicable. The authors take responsibility for the final text, citations, claims, and reported results. Generative AI was not used to generate synthetic datasets or to prove mathematical claims.

\bibliography{iclr2027_conference}
\bibliographystyle{iclr2027_conference}

\appendix
\vspace{-5pt}
\section{Supplementary Package Overview}
\vspace{-5pt}
This appendix documents the supplementary implementation package accompanying CALIBUDGET. It specifies the package boundary, deterministic allocation and integrity contracts, CPU-checkable validation procedure, and the relation between implementation components and the evidence reported in the main paper. It also consolidates claim boundaries and records separately completed audits without pooling them into the confirmatory three-seed results. The appendix introduces no new method, experimental setting, or headline claim.
\vspace{-5pt}
\section{Scope and Relationship to the Main Paper}
\vspace{-5pt}
The supplied archive is designed to make the allocation and audit layer inspectable without requiring model weights, datasets, network access, or GPU execution. It contains two complementary forms of material. First, the compact package under \path{src/calibudget/} implements dependency-light allocation, record-validation, tokenization-contract, sketching, and file-binding primitives used by synthetic tests. Second, \path{source/} contains sanitized source snapshots of the training, allocation, proxy-contract, and external-evaluation implementations used in the study. These snapshots expose the relevant control flow and contracts, but deliberately omit private artifact roots, launch infrastructure, checkpoints, generated outputs, and machine-specific configuration.

The package therefore supports two levels of inspection. The default checks verify deterministic behavior, fail-closed validation, and file integrity on a CPU-only machine. Full experiment reproduction additionally requires the public model and dataset dependencies, the registered training environment, and locally supplied paths. This distinction prevents the lightweight verification package from being mistaken for a self-contained distribution of model weights or experimental outputs.

\begin{center}
\footnotesize
\setlength{\tabcolsep}{5.0pt}
\renewcommand{\arraystretch}{1.08}
\begin{tabularx}{\textwidth}{@{}>{\raggedright\arraybackslash}p{3.6cm}>{\raggedright\arraybackslash}X>{\raggedright\arraybackslash}p{4.2cm}@{}}
\toprule
Package component & Purpose & Verification boundary \\
\midrule
\path{src/calibudget/allocation.py} & Deterministic floor--residual allocation, capacity-aware largest-remainder rounding, and selected-index hashing. & Model- and dataset-independent synthetic inputs. \\
\path{src/calibudget/proxy.py} & Canonical subgroup-score record validation and stable four-shard aggregation. & Validates structure and ordering without exposing experiment-specific score values. \\
\path{src/calibudget/data.py} & Record normalization, normalized prompt hashing, and pairwise-disjoint role checks. & Dataset-agnostic integrity contracts. \\
\path{src/calibudget/modeling.py} & Response-only tokenization-length contract with caller-supplied tokenizer. & No model download or model identity is required. \\
\path{src/calibudget/sketch.py} & Deterministic CountSketch for finite one-dimensional vectors. & Synthetic numeric arrays only. \\
\path{src/calibudget/audit.py} & Strict JSON-object loading, SHA-256 calculation, and regular-file binding checks. & Local file integrity and provenance checks. \\
\path{source/} & Sanitized snapshots of training, allocation, proxy-contract, and modern-task evaluation code. & Source inspection and compilation; generated outputs and launch infrastructure are excluded. \\
\path{tests/} & Seven synthetic tests covering allocation, proxy records, role separation, sketch determinism, and file binding. & Runs without network access, model loading, or datasets. \\
\bottomrule
\end{tabularx}
\captionof{table}{Map of the supplementary package. The compact modules expose the contracts exercised by the default tests, while the sanitized snapshots preserve additional implementation context.}
\label{tab:package-map}
\end{center}

\section{Deterministic Allocation Contract}

\subsection{Floor--Residual Decomposition}

For source capacities $N_k$, total source budget $B$, and floor $f$, the compact implementation first assigns
\begin{equation}
q_k^{\mathrm{floor}}=\min(N_k,f), \qquad
R=B-\sum_{k=1}^{K}q_k^{\mathrm{floor}}.
\end{equation}
It rejects negative budgets or capacities, mismatched source keys, an infeasible floor, and any budget exceeding total capacity. For active sources with residual capacity $c_k=N_k-q_k^{\mathrm{floor}}>0$, nonnegative utilities are normalized over the active set. Real-valued residual shares are floored, capped by $c_k$, and completed by largest-remainder redistribution until the residual sum is exactly $R$. The returned quotas satisfy
\begin{equation}
\min(N_k,f)\le q_k\le N_k,
\qquad \sum_{k=1}^{K}q_k=B.
\end{equation}

The production allocation snapshot uses the frozen defaults $B=10{,}000$, $f=1{,}150$, $(\alpha,\beta,\gamma)=(1,0.5,1)$, and seed 42 unless overridden. It computes reliability as $(1+\sigma_k)^{-1}$ and post-floor availability from the square root of remaining capacity. The compact allocator intentionally accepts already constructed utility values; this separation permits the quota mechanism to be tested independently of model-based score estimation.

\subsection{Stable Tie Breaking and Selection Identity}

Every label traversal is lexically sorted, and equal fractional remainders are resolved by the source label. Consequently, the same capacities, utilities, budget, and floor produce the same quota dictionary across Python processes and platforms. The synthetic allocation test uses capacities $(3,10,10)$, utilities $(1,2,3)$, $B=12$, and $f=2$, and requires the exact output $(3,4,5)$.

Selected-example identity is represented separately from quota identity. The helper \path{canonical_selected_indices_sha256} rejects duplicate indices and hashes the ordered sequence. Order sensitivity is deliberate: two runs selecting the same set in a different stored order do not silently share an identifier. The production snapshot additionally verifies that selected examples do not overlap calibration indices and that the selected count equals the requested source budget.

\section{Data, Proxy, and Audit Contracts}

\subsection{Record and Role Integrity}

The record normalizer accepts common field aliases such as instruction/prompt/question and response/output/answer, collapses instruction whitespace, strips the response, and rejects missing instruction--response pairs. When no explicit record ID is present, a SHA-256 ID is derived from the normalized instruction and response. Prompt hashing is case-insensitive after whitespace normalization.

Role indices are required to be nonempty, unique within each role, pairwise disjoint across roles, and named by lowercase identifiers. This contract is used to prevent accidental crossing among candidate, calibration, validation, or other registered roles. It is a structural safeguard rather than a claim that the archive contains the original datasets.

\subsection{Proxy-Score and Shard Validation}

The proxy contract freezes the eight-source order as ARC-Challenge, ARC-Easy, BoolQ, HellaSwag, OpenBookQA, PIQA, Social IQa, and WinoGrande. Each score record must include a canonical position, record ID, all eight finite subgroup means, and a finite aggregate score equal to the recomputed maximum subgroup mean. Aggregation requires exactly four canonical shards. A record at position $i$ must appear in shard $i\bmod 4$; missing, duplicated, out-of-range, or cross-shard positions are rejected. After validation, records are ordered by descending score and then by candidate ID, yielding a stable selected-ID sequence.

These checks are intentionally fail-closed. They do not publish the experiment-specific proxy values, but they make malformed score records, subgroup-order drift, incomplete unions, and unstable tie breaking detectable before selection.

\subsection{Tokenization, Sketching, and File Binding}

The response-only tokenization helper constructs a fixed instruction/response template, applies truncation with a caller-supplied tokenizer, and reports both total length and the number of response tokens left unmasked. It rejects nonpositive maximum lengths and examples whose response has no surviving token after truncation.

The CountSketch implementation uses a deterministic BLAKE2b-derived bucket and sign for each vector position, processes long vectors in blocks, and rejects nonfinite values. Audit helpers require JSON roots to be objects and bind files by regular-file status, exact byte count, and SHA-256 digest; symbolic links are rejected for bound artifacts.

\section{CPU-Checkable Verification Procedure}

The archive includes a PowerShell entry point and can be checked equivalently on a POSIX shell. From the archive root, the following commands compile the compact modules, sanitized snapshots, and tests, then execute the synthetic suite:

\begin{quote}
\small\ttfamily
PYTHONPATH=src python -m compileall -q src source tests\\
PYTHONPATH=src python -m pytest -q tests
\end{quote}

The distributed suite contains seven tests. They verify: (i) exact budget and capacity safety of floor--residual allocation; (ii) order-sensitive selected-index hashes; (iii) canonical four-shard membership and stable top-$B$ selection; (iv) rejection of subgroup-order drift; (v) record normalization and role disjointness; (vi) deterministic sketching and rejection of nonfinite vectors; and (vii) JSON/file-binding integrity. The archive also provides \path{SHA256SUMS.tsv}, which binds 26 distributed files by relative path, byte count, and SHA-256 digest. During appendix preparation, all 26 bindings matched and all seven synthetic tests passed in a clean CPU-only check.

\section{Reproduction Boundary and Local Configuration}

The local path template exposes placeholders for model, data, and artifact roots and records the experimental defaults needed by the allocation layer: seed 42, commonsense budget 10,000, mathematical companion budget 100,000, and floor 1,150. Machine-specific paths are intentionally absent. The archive also excludes model weights, datasets, credentials, network launchers, SSH helpers, checkpoints, logs, and generated experiment outputs.

The compact verification modules require NumPy, PyTorch, and pytest as listed in \path{requirements.txt}; the default tests do not import optional model-training dependencies. Executing the sanitized training and evaluation snapshots requires the additional scientific and model-training environment used for the registered experiments. The code package therefore supports source inspection and contract validation directly, while end-to-end reruns depend on separately obtained public resources and local compute.

\section{Claim Boundaries and Additional Audits}
\label{app:claim-boundaries}
Table~\ref{tab:claim-boundary} pairs each evidence layer with the strongest conclusion supported by the current results and with a stronger conclusion that remains unestablished. This separation is important because the confirmatory study uses three paired seeds, whereas transfer and diagnostic analyses are intended to delimit scope rather than select a new method.

\begin{center}
\footnotesize
\setlength{\tabcolsep}{4.2pt}
\renewcommand{\arraystretch}{1.06}
\begin{tabularx}{\textwidth}{@{}>{\raggedright\arraybackslash}p{2.85cm}>{\raggedright\arraybackslash}X>{\raggedright\arraybackslash}X@{}}
\toprule
Evidence layer & What the evidence supports & What it does not establish \\
\midrule
Main paired study & Consistent gains on CommonAvg, FragileAvg, and MacroAvg under matched LLaMA-2 budgets and LoRA+ controls. & Uniform improvement, large-sample significance, or absence of a MathAvg trade-off. \\
Reliability ablation & The complete utility is consistently stronger than removing $\rho_k$ on source-balanced metrics in the three observed seeds. & Independent causal effects among interacting utility terms. \\
Quota geometry & Small, structured source reallocations can accompany measurable aggregate changes. & A monotonic quota--accuracy response for individual tasks. \\
Qwen2.5 audit & Directionally similar descriptive gains under an independently recalibrated second backbone. & Backbone-invariant superiority; all reported bootstrap intervals cross zero. \\
LLaMA-3.1 and modern panel & Transferred LLaMA-3.1 allocations are near parity, and the broader modern-task panel shows no reliable gain. & Reliable gains under transferred quotas, on code tasks, or with full-parameter tuning. \\
Floor and split audits & Quota rankings are stable across calibration splits, while the floor operating point remains consequential. & A universal floor value or robustness to arbitrary source partitions. \\
\bottomrule
\end{tabularx}
\captionof{table}{Claim boundary. Each evidence layer is paired with the conclusion supported by the current results and with a stronger conclusion that those results do not establish.}
\label{tab:claim-boundary}
\end{center}

\subsection{Separately Completed Qwen2.5 Reconstruction}

Table~\ref{tab:qwen-reconstruction} records a separately completed Qwen2.5-7B seed-42 reconstruction. It was run under matched source budget, training, and evaluation controls. Because it is a single-seed consistency audit, it is not pooled with the three-seed Qwen2.5 summary in the main paper.

\begin{center}
\footnotesize
\setlength{\tabcolsep}{3.7pt}
\renewcommand{\arraystretch}{1.06}
\begin{tabular}{@{}lrrrrr@{}}
\toprule
Method & Math & Common & Fragile & Macro & Overall \\
\midrule
\rowcolor{CalibPaleGray}
Val-error+floor & 65.38 & 84.55 & 88.04 & 80.71 & 74.96 \\
\rowcolor{CalibPaleBlue}
\textbf{CALIBUDGET} & 65.11 & 84.93 & 88.54 & 80.97 & 75.02 \\
\midrule
$\Delta$ (pp) & $-0.26$ & $+0.39$ & $+0.50$ & $+0.26$ & $+0.06$ \\
\bottomrule
\end{tabular}
\captionof{table}{Separately audited Qwen2.5-7B seed-42 reconstruction (accuracy, \%). Differences are CALIBUDGET minus Val-error+floor and are computed before display rounding.}
\label{tab:qwen-reconstruction}
\end{center}

\subsection{Completed Component Check}

A separately registered LLaMA-2 component audit had completed seeds 42--43 at the evidence cutoff. Need+reliability exceeded need+availability by $+0.20$ pp on CommonAvg, $+0.68$ pp on FragileAvg, $+0.15$ pp on MacroAvg, and $+0.08$ pp on Overall, with a $-0.03$ pp MathAvg difference. The direction is consistent with the reliability interpretation in the main paper, but seed 44 and the complete matched component grid were unfinished. We therefore treat this result as provisional convergent evidence, not as a final ablation or a basis for revising the frozen method.

\paragraph{Practical audit checklist.}
{\small An independent rerun should retain source definitions and counts, calibration-role indices, need and bootstrap-stability statistics, floor, exponents, total budget, residual shares, final quotas, selected-index hash, random seed, parser/scorer versions, and per-task outputs. Execution should fail on role overlap, file-size or hash mismatch, subgroup-order drift, incomplete shards, capacity violation, or an incorrect final count. These records make the allocation decision auditable independently of downstream model training.\par}

\end{document}